\def\year{2019}\relax
\documentclass[letterpaper]{article}
\usepackage{aaai19}
\usepackage{times}
\usepackage{helvet}
\usepackage{courier}
\usepackage{url}
\usepackage{graphicx}
\usepackage{multirow}
\usepackage{amsfonts}
\usepackage{algorithm}
\usepackage{algorithmic}
\usepackage{booktabs}
\usepackage{subfigure}

\begin{document}
\title{Unsupervised Bilingual Lexicon Induction from Mono-lingual Multimodal Data}
\author{Shizhe Chen,\textsuperscript{\rm 1}
	Qin Jin,\textsuperscript{\rm 1}\thanks{Qin Jin is the corresponding author.}
	Alexander Hauptmann\textsuperscript{\rm 2}\\
	\textsuperscript{\rm 1} School of Information, Renmin University of China, Beijing, China\\
	\textsuperscript{\rm 2} Language Technology Institude, Carnegie Mellon University, Pittsburgh, USA \\
	\{cszhe1, qjin\}@ruc.edu.cn, alex@cs.cmu.edu
}
\maketitle
\begin{abstract}
Bilingual lexicon induction, translating words from the source language to the target language,
is a long-standing natural language processing task. 
Recent endeavors prove that it is promising to employ images as pivot to learn the lexicon induction without reliance on parallel corpora. 
However, these vision-based approaches simply associate words with entire images, which are constrained to translate concrete words and require object-centered images. 
We humans can understand words better when they are within a sentence with context. 
Therefore, in this paper, we propose to utilize images and their associated captions to address the limitations of previous approaches.
We propose a multi-lingual caption model trained with different mono-lingual multimodal data to map words in different languages into joint spaces.
Two types of word representation are induced from the multi-lingual caption model: linguistic features and localized visual features.
The linguistic feature is learned from the sentence contexts with visual semantic constraints, which is beneficial to learn translation for words that are less visual-relevant.
The localized visual feature is attended to the region in the image that correlates to the word, so that it alleviates the image restriction for salient visual representation. 
The two types of features are complementary for word translation.
Experimental results on multiple language pairs demonstrate the effectiveness of our proposed method, which substantially outperforms previous vision-based approaches without using any parallel sentences or supervision of seed word pairs.
\end{abstract}

\section{Introduction}

The bilingual lexicon induction task aims to automatically build word translation dictionaries across different languages, which is beneficial for various natural language processing tasks such as cross-lingual information retrieval \cite{lavrenko2002cross}, multi-lingual sentiment analysis \cite{denecke2008using}, machine translation \cite{och2003systematic} and so on.
Although building bilingual lexicon has achieved success with parallel sentences in resource-rich languages \cite{och2003systematic}, the parallel data is insufficient or even unavailable especially for resource-scarce languages and it is expensive to collect.
On the contrary, there are abundant multimodal mono-lingual data on the Internet, such as images and their associated tags and descriptions, which motivates researchers to induce bilingual lexicon from these non-parallel data without supervision.

\begin{figure} \centering
	\subfigure[The previous vision-based approach: a word is represented by global features extracted from retrieved images. It requires object-centered images and is unreliable for non-concrete words.] { \centering
		\label{fig:intro_previous_method}
		\includegraphics[width=0.95\linewidth]{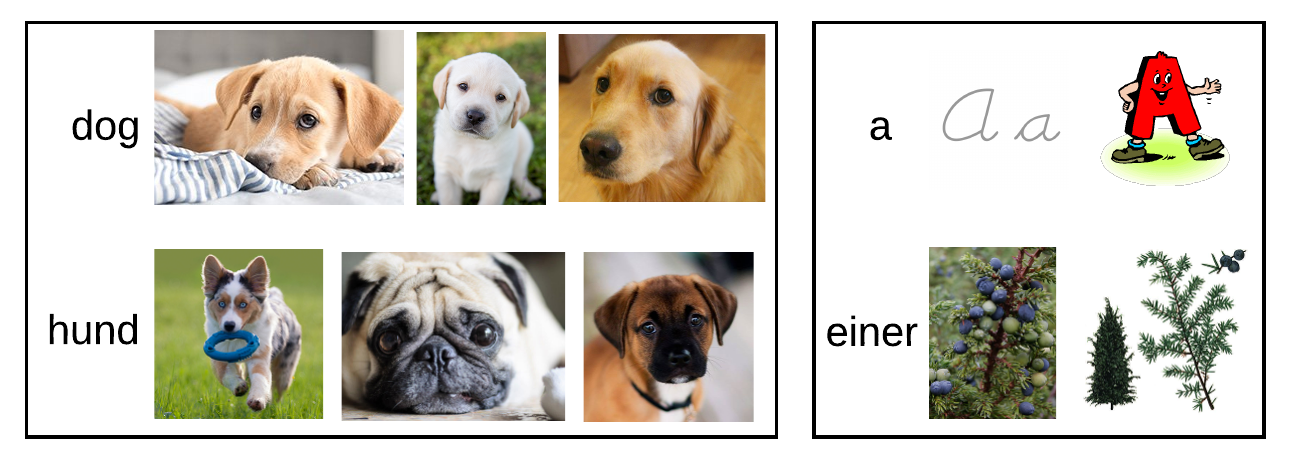}
	} 
	\subfigure[Our proposed approach: the word representation is learned from both sentence contexts and visual localization.] { \centering
		\label{fig:intro_proposed_method}
		\includegraphics[width=0.95\linewidth]{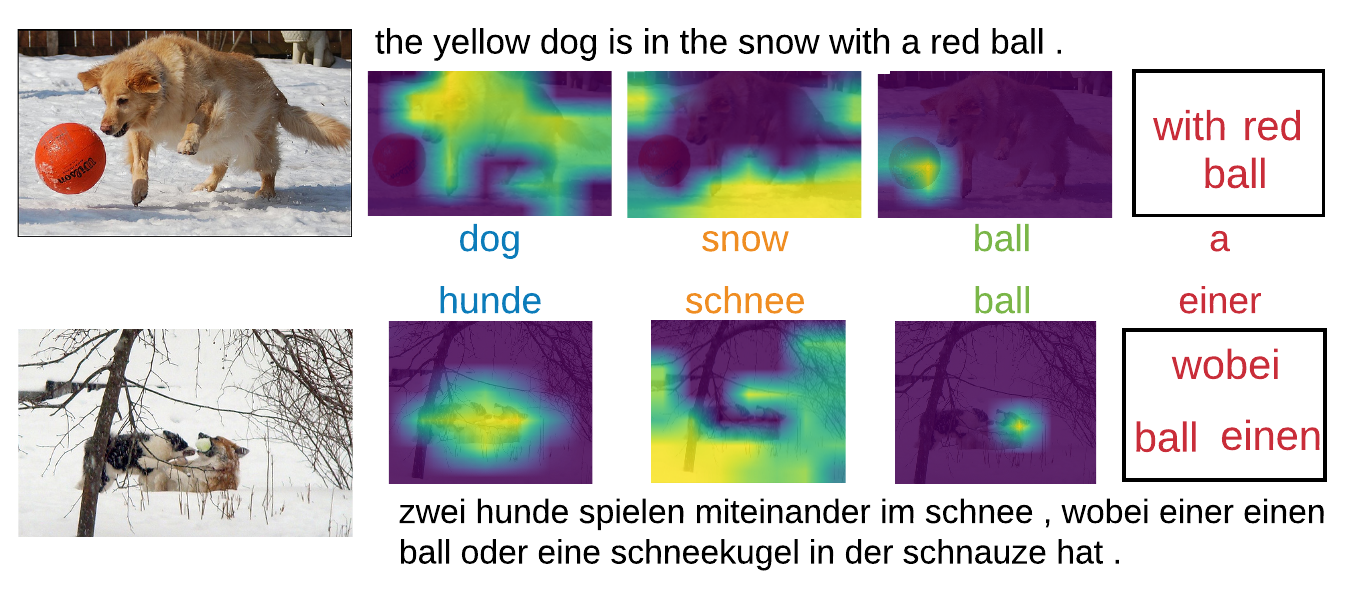}
	}
	\caption{Comparison of previous vision-based approaches and our proposed approach for bilingual lexicon induction. Best viewed in color.}
	\label{fig:intro_motivation}
\end{figure}

There are mainly two types of mono-lingual approaches to build bilingual dictionaries in recent works.
The first is purely text-based, which explores the structure similarity between different linguistic space.
The most popular approach among them is to linearly map source word embedding into the target word embedding space \cite{mikolov2013exploiting,conneau2017word}.
The second type utilizes vision as bridge to connect different languages \cite{bergsma2011learning,kiela2015visual,hewitt2018learning}.
It assumes that words correlating to similar images should share similar semantic meanings. 
So previous vision-based methods search images with multi-lingual words and translate words according to similarities of visual features extracted from the corresponding images.
It has been proved that the visual-grounded word representation improves the semantic quality of the words \cite{silberer2012grounded}.

However, previous vision-based methods suffer from two limitations for bilingual lexicon induction.
Firstly, the accurate translation performance is confined to concrete visual-relevant words such as nouns and adjectives as shown in Figure~\ref{fig:intro_previous_method}.
For words without high-quality visual groundings, previous methods would generate poor translations \cite{hewitt2018learning}.
Secondly, previous works extract visual features from the whole image to represent words and thus require object-centered images in order to obtain reliable visual groundings. 
However, common images usually contain multiple objects or scenes, and the word might only be grounded to part of the image, therefore the global visual features will be quite noisy to represent the word.

In this paper, we address the two limitations via learning from mono-lingual multimodal data with both sentence and visual context (e.g., image and caption data) to induce bilingual lexicon.
Such multimodal data is also easily obtained for different languages on the Internet \cite{sharma2018conceptual}.
We propose a multi-lingual image caption model trained on multiple mono-lingual image caption data, which is able to induce two types of word representations for different languages in the joint space.
The first is the linguistic feature learned from the sentence context with visual semantic constraints, so that it is able to generate more accurate translations for words that are less visual-relevant.
The second is the localized visual feature which attends to the local region of the object or scene in the image for the corresponding word, so that the visual representation of words will be more salient than previous global visual features.
The two representations are complementary and can be combined to induce better bilingual word translation.

We carry out experiments on multiple language pairs including German-English,  French-English, and Japanese-English.
The experimental results show that the proposed multi-lingual caption model not only achieves better caption performance than independent mono-lingual models for data-scarce languages, but also can induce the two types of features, linguistic and visual features, for different languages in joint spaces.
Our proposed method consistently outperforms previous state-of-the-art vision-based bilingual word induction approaches on different languages.
The contributions of this paper are as follows:
\begin{itemize}
	\item To the best of our knowledge, we are the first to explore images associated with sentences for bilingual lexicon induction, which mitigates two main limitations of previous vision-based approaches.
	\item We propose a multi-lingual caption model to induce both the linguistic and localized visual features for multi-lingual words in joint spaces. The linguistic and visual features are complementary to enhance word translation.
	\item Extensive experiments on different language pairs demonstrate the effectiveness of our approach, which achieves significant improvements over the state-of-the-art vision-based methods in all part-of-speech classes.
\end{itemize}

\section{Related Work}

The early works for bilingual lexicon induction require parallel data in different languages.
\cite{och2003systematic} systematically investigates various word alignment methods with parallel texts to induce bilingual lexicon.
However, the parallel data is scarce or even unavailable for low-resource languages. 
Therefore, methods with less dependency on the availability of parallel corpora are highly desired.

There are mainly two types of mono-lingual approaches for bilingual lexicon induction: text-based and vision-based methods.
The text-based methods purely exploit the linguistic information to translate words.
The initiative works \cite{fung1998ir,rapp1999automatic} utilize word co-occurrences in different languages as clue for word alignment.
With the improvement in word representation based on deep learning, \cite{mikolov2013exploiting} finds the structure similarity of the deep-learned word embeddings in different languages, and employs a parallel vocabulary to learn a linear mapping from the source to target word embeddings.
\cite{xing2015normalized} improves the translation performance via adding an orthogonality constraint to the mapping.
\cite{zhang2017bilingual} further introduces a matching mechanism to induce bilingual lexicon with fewer seeds.
However, these models require seed lexicon as the start-point to train the bilingual mapping.
Recently, \cite{conneau2017word} proposes an adversarial learning approach to learn the joint bilingual embedding space without any seed lexicon.

The vision-based methods exploit images to connect different languages, which assume that words corresponding to similar images are semantically alike.
\cite{bergsma2011learning} collects images with labeled words in different languages to learn word translation with image as pivot.
\cite{kiela2015visual} improves the visual-based word translation performance via using more powerful visual representations: the CNN-based \cite{krizhevsky2012imagenet} features. 
The above works mainly focus on the translation of nouns and are limited in the number of collected languages.
The recent work \cite{hewitt2018learning} constructs the current largest (with respect to the number of language pairs and types of part-of-speech) multimodal word translation dataset, MMID.
They show that concrete words are easiest for vision-based translation methods while others are much less accurate.
In our work, we alleviate the limitations of previous vision-based methods via exploring images and their captions rather than images with unstructured tags to connect different languages.

Image captioning has received more and more research attentions.
Most image caption works focus on the English caption generation \cite{vinyals2017show,xu2015show}, while there are limited works considering generating multi-lingual captions.
The recent WMT workshop \cite{elliott2017findings} has proposed a subtask of multi-lingual caption generation, where different strategies such as multi-task captioning and source-to-target translation followed by captioning have been proposed to generate captions in target languages.
Our work proposes a multi-lingual image caption model that shares part of the parameters across different languages in order to benefit each other.

\section{Unsupervised Bilingual Lexicon Induction}
Our goal is to induce bilingual lexicon without supervision of parallel sentences or seed word pairs, purely based on the mono-lingual image caption data.
In the following, we introduce the multi-lingual image caption model whose objectives for bilingual lexicon induction are two folds: 1) explicitly build multi-lingual word embeddings in the joint linguistic space; 2) implicitly extract the localized visual features for each word in the shared visual space. 
The former encodes linguistic information of words while the latter encodes the visual-grounded information, which are complementary for bilingual lexicon induction.

\subsection{Multi-lingual Image Caption Model}
Suppose we have mono-lingual image caption datasets $D_{v, x}=\{\langle v^{(i)}, x^{(i)} \rangle\}_{i=1}^{M}$ in the source language and $D_{v, y}=\{\langle v^{(i)}, y^{(i)} \rangle\}_{i=1}^{N}$ in the target language.
The images $v$ in $D_{v, x}$ and $D_{v, y}$ do not necessarily overlap, but cover overlapped object or scene classes which is the basic assumption of vision-based methods.
For notation simplicity, we omit the superscript $i$ for the data sample.
Each image caption $x$ and $y$ is composed of word sequences $\{x_1, ..., x_{T_x}\}$ and $\{y_1, ..., y_{T_y}\}$ respectively, where $T_{*}$ is the sentence length.

\begin{figure}
	\begin{center}
		\includegraphics[width=1\linewidth]{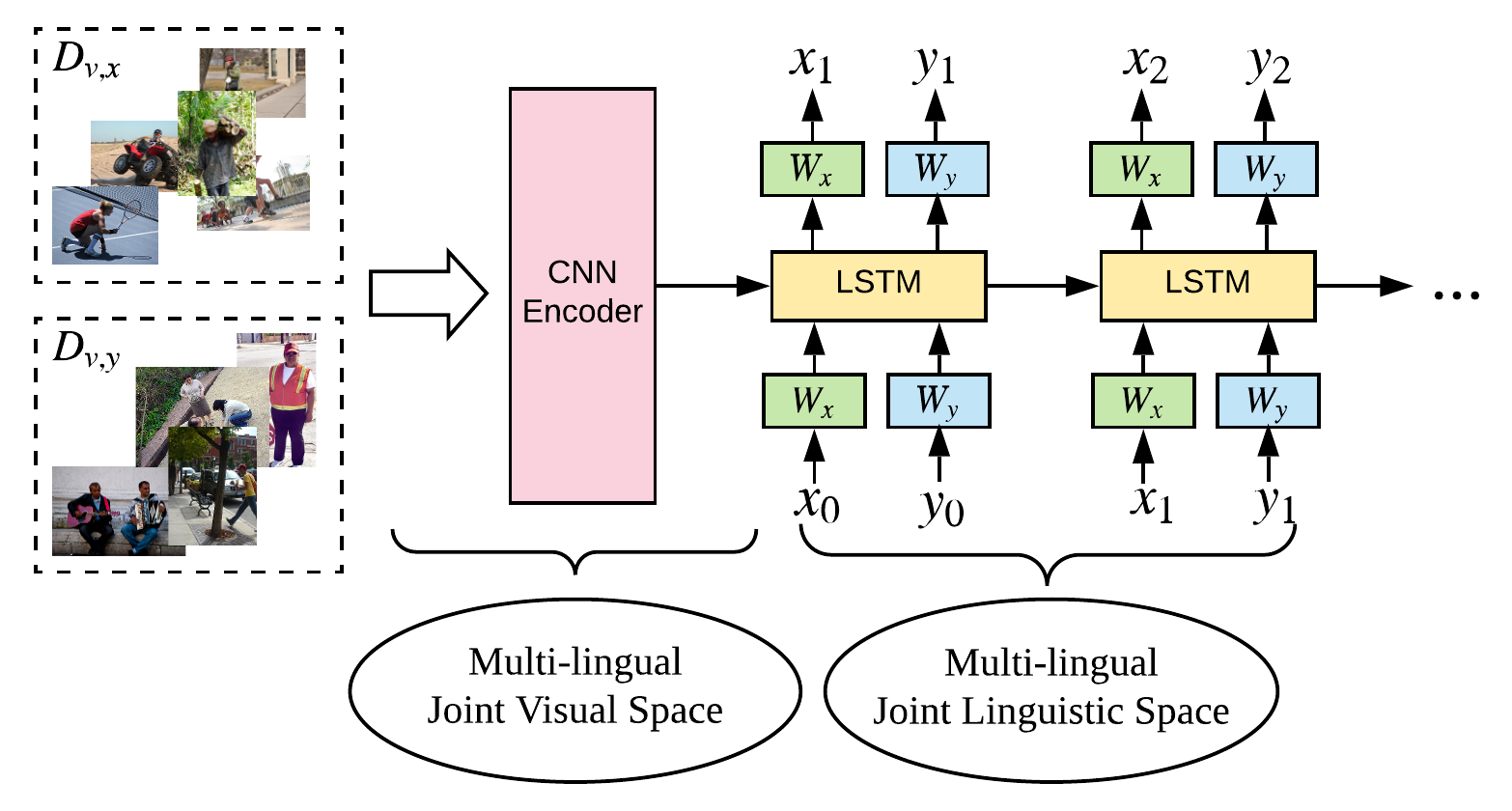}
	\end{center}
	\caption{Multi-lingual image caption model. The source and target language caption models share the same image encoder and language decoder, which enforce the word embeddings of different languages to project in the same space.}
	\label{fig:caption_model}
\end{figure}

The proposed multi-lingual image caption model aims to generate sentences in different languages to describe the image content, which connects the vision and multi-lingual sentences.
Figure~\ref{fig:caption_model} illustrates the framework of the caption model, which consists of three parts: the image encoder, word embedding module and language decoder.

The image encoder encodes the image into the shared visual space.
We apply the Resnet152 \cite{he2016deep} as our encoder $f_e$, which produces $K$ vectors corresponding to different spatial locations in the image:
\begin{equation}
a = \{a_1, ..., a_{K}\} = f_{e}(v; \theta_{e})
\end{equation}
where  $a_{k} \in \mathbb{R}^{D}$.
The parameter $\theta_{e}$ of the encoder is shared for different languages in order to encode all the images in the same visual space.

The word embedding module maps the one-hot word representation in each language into low-dimensional distributional embeddings:
\begin{equation}
w^{x}_{t} = W_x x_t; \quad
w^{y}_{t} = W_y y_t
\end{equation}
where $W_{x} \in \mathbb{R}^{D \times N_x}$ and $W_{y} \in \mathbb{R}^{D \times N_y}$ is the word embedding matrix for the source and target languages respectively. $N_x$ and $N_y$ are the vocabulary size of the two languages.

The decoder then generates word step by step conditioning on the encoded image feature and previous generated words. 
The probability of generating $x_t$ in the source language is as follows:
\begin{equation}
\label{eq:word_pred}
\mathrm{P}(x_t|a, x_{<t}) = f_{d}(h^x_t; \theta_{d}) = \mathrm{softmax}(h_t^x W_x)
\end{equation}
where $h_t^x$ is the hidden state of the decoder at step $t$, which is functioned by LSTM \cite{hochreiter1997long}:
\begin{equation}
h_t^x = \mathrm{LSTM}([w^{x}_{t-1}; c^x_t], h_{t-1}^x)
\end{equation}
The $c^x_t$ is the dynamically located contextual image feature to generate word $x_t$ via attention mechanism, which is the weighted sum of $a$ computed by 
\begin{equation}
\label{eq:attention_ft}
c^x_t =\sum_{k=1}^{K} \alpha_{t, k} a_{k}
\end{equation}
\begin{equation}
\label{eq:attention_alpha}
\alpha_{t, k} = \frac{\mathrm{exp} (f_{a}(h_{t-1}^x, a_{k}))} {\sum_{k=1}^{K} \mathrm{exp}(f_{a}(h_{t-1}^x, a_{k}))}
\end{equation}
where $f_a$ is a fully connected neural network. The parameter $\theta_{d}$ in the decoder includes all the weights in the LSTM and the attention network $f_a$.

Similarly,  $\mathrm{P}(y_t|a, y_{<t})$ is the probability of generating $y_t$ in the target language, which shares $\theta_{d}$ with the source language.
By sharing the same parameters across different languages in the encoder and decoder, both the visual features and the learned word embeddings for different languages are enforced to project in a joint semantic space.
To be noted, the proposed multi-lingual parameter sharing strategy is not constrained to the presented image captioning model, but can be applied in various image captioning models such as show-tell model \cite{vinyals2017show} and so on.

We use maximum likelihood as objective function to train the multi-lingual caption model, which maximizes the log-probability of the ground-truth captions:
\begin{equation}
L = \sum_{t=1}^{T_x} \mathrm{log} \mathrm{P}(x_t|a, x_{<t}) + \sum_{t=1}^{T_y} \mathrm{log} \mathrm{P}(y_t|a, y_{<t})
\end{equation}

\subsection{Visual-guided Word Representation}
The proposed multi-lingual caption model can induce similarities of words in different languages from two aspects: the linguistic similarity and the visual similarity.
In the following, we discuss the two types of similarity and then construct the source and target word representations.

The \textbf{linguistic similarity} is reflected from the learned word embeddings $W_x$ and $W_y$ in the multi-lingual caption model.
As shown in previous works \cite{mikolov2013linguistic}, word embeddings learned from the language contexts can capture syntactic and semantic regularities in the language.
However, if the word embeddings of different languages are trained independently, they are not in the same linguistic space and we cannot compute similarities directly.
In our multi-lingual caption model, since images in $D_{v, x}$ and $D_{v, y}$ share the same visual space, the features of sentence $x$ and $y$ belonging to similar images are bound to be close in the same space with the visual constraints.
Meanwhile, the language decoder is also shared, which enforces the word embeddings across languages into the same semantic space in order to generate similar sentence features.
Therefore, $W_x$ and $W_y$ not only encode the linguistic information of different languages but also share the embedding space which enables direct cross-lingual similarity comparison.
We refer the linguistic features of source and target words $x_n$ and $y_n$ as $l^x_n$ and $l^y_n$ respectively.

For the \textbf{visual similarity}, the multi-lingual caption model locates the image region to generate each word base on the spatial attention in Eq~(\ref{eq:attention_alpha}), which can be used to calculate the localized visual representation of the word.
However, since the attention is computed before word generation, the localization performance can be less accurate.
It also cannot be generalized to image captioning models without spatial attention.
Therefore, inspired by \cite{zeiler2014visualizing}, where they occlude over regions of the image to observe the change of classification probabilities, we feed different parts of the image to the caption model and investigate the probability changes for each word in the sentence.
Algorithm~\ref{alg:attention_refine} presents the procedure of word localization and the grounded visual feature generation.
Please note that such visual-grounding is learned unsupervisedly from the image caption data.
Therefore, every word can be represented as a set of grounded visual features (the set size equals to the word occurrence number in the dataset).
We refer the localized visual feature set for source word $x_n$ as $\{i^x_{n, 1}, ..., i^x_{n, C_n^x}\}$, for target word $y_n$ as $\{i^y_{n, 1}, ..., i^y_{n, C_n^y}\}$.

\begin{algorithm}
	\caption{Generating localized visual features.}
	\label{alg:attention_refine}
	\begin{algorithmic}
		\REQUIRE {Encoded image features $a=\{a_1, ..., a_K\}$, sentence $x=\{x_1, ..., x_T\}$.}
		\ENSURE  Localized visual features for each word
		\FOR {$t \in [1, T]$}
		\FOR{each $a_k \in a$}
		\STATE compute $\mathrm{P}_{t, k}=\mathrm{P}(x_t|a_k, x_{<t})$ according to Eq~(\ref{eq:word_pred})
		\ENDFOR
		\STATE $\hat{\alpha}_{t, k}=\mathrm{P}_{t, k} / (\sum_k \mathrm{P}_{t, k})$
		\STATE $\hat{c}_{t} = \sum_k \hat{\alpha}_{t, k} a_k$
		\ENDFOR
		\RETURN  $\{\hat{c}_1, ..., \hat{c}_T\}$
	\end{algorithmic}
\end{algorithm}

\subsection{Word Translation Prediction}
Since the word representations of the source and target language are in the same space, we could directly compute the similarities across languages.
We apply l2-normalization on the word representations and measure with the cosine similarity. 
For linguistic features, the similarity is measured as:
\begin{equation}
s_l(x_n, y_m) = l^x_n \cdot l^y_m
\end{equation}

However, there are a set of visual features associated with one word, so the visual similarity measurement between two words is required to take two sets of visual features as input.
We aggregate the visual features in a single representation and then compute cosine similarity instead of point-wise similarities among two sets:
\begin{equation}
s_{i}(x_n, y_m) = (\frac{1}{C_n^x} \sum_{j=1}^{C_n^x} i^x_{n, j}) \cdot (\frac{1}{C_m^y} \sum_{j=1}^{C_m^y} i^y_{m, j})
\end{equation}
The reasons for performing aggregation are two folds.
Firstly, the number of visual features is proportional to the word occurrence in our approach instead of fixed numbers as in \cite{kiela2015visual,hewitt2018learning}. So the computation cost for frequent words are much higher.
Secondly, the aggregation helps to reduce noise, which is especially important for abstract words. The abstract words such as `event' are more visually diverse, but the overall styles of multiple images can reflect its visual semantics.

Due to the complementary characteristics of the two features, we combine them to predict the word translation. The translated word for $x_n$ is 
\begin{equation}
y_{*} = \mathrm{arg max}_{y_m} (s_l (x_n, y_m) + s_i (x_n, y_m))
\end{equation}

\section{Experiments}

\begin{table}
	\caption{Statistics of image caption datasets.}
	\label{tab:caption_datasets}
	\begin{tabular}{c|c|ccc} \toprule
		dataset & lang & \#images & \#captions & \#words \\ \midrule
		\multirow{3}{*}{Multi30k} & English & 14,500 & 72,500 & 5,492 \\
		& German & 14,500 & 72,500 & 5,535 \\
		& French & 14,500 & 14,500 & 3,318 \\ \midrule
		\multirow{2}{*}{\begin{tabular}[c]{@{}c@{}}COCO\\ + STAIR\end{tabular}} & English & 56,644 & 283,378 & 7,491 \\
		& Japanese & 56,643 & 283,215 & 9,331 \\ \bottomrule
	\end{tabular}
\end{table}
\begin{table*} \centering
	\caption{Image captioning performance of different languages on the Multi30k dataset.}
	\label{tab:multi30k_caption_results}
	\begin{tabular}{c|ccc|ccc|ccc} \toprule
		& \multicolumn{3}{c|}{English} & \multicolumn{3}{c|}{German} & \multicolumn{3}{c}{French} \\
		& BLEU4 & METEOR & CIDEr & BLEU4 & METEOR & CIDEr & BLEU4 & METEOR & CIDEr \\ \midrule
		mono-lingual model & \textbf{23.59} & 20.62 & 50.19 & 16.02 & 18.79 & 44.24 & 6.39 & 13.60 & 47.45 \\
		multi-lingual model & 23.39 & \textbf{20.86} & \textbf{51.04} & \textbf{16.47} & \textbf{18.82} & \textbf{44.75} & \textbf{7.15} & \textbf{13.77} & \textbf{50.67}  \\ \bottomrule
	\end{tabular}
\end{table*}

\subsection{Datasets}
For image captioning, we utilize the multi30k  \cite{elliott2016multi30k}, COCO \cite{chen2015microsoft} and STAIR \cite{yoshikawa2017stair} datasets.
The multi30k dataset contains 30k images and annotations under two tasks. 
In task 1, each image is annotated with one English description which is then translated into German and French.
In task 2, the image is independently annotated with 5 descriptions in English and German respectively.
For German and English languages, we utilize annotations in task 2.
For the French language, we can only employ French descriptions in task 1, so the training size for French is less than the other two languages.
The COCO and STAIR datasets contain the same image set but are independently annotated in English and Japanese.
Since the images in the wild for different languages might not overlap, we randomly split the image set into two disjoint parts of equal size.
The images in each part only contain the mono-lingual captions.
We use Moses SMT Toolkit to tokenize sentences and select words occurring more than five times in our vocabulary for each language.
Table~\ref{tab:caption_datasets} summarizes the statistics of caption datasets. 

For bilingual lexicon induction, we use two visual datasets: BERGSMA and MMID.
The BERGSMA dataset \cite{bergsma2011learning} consists of 500 German-English word translation pairs.
Each word is associated with no more than 20 images.
The words in BERGSMA dataset are all nouns.
The MMID dataset \cite{hewitt2018learning} covers a larger variety of words and languages, including 9,808 German-English pairs and 9,887 French-English pairs.
The source word can be mapped to multiple target words in their dictionary.
Each word is associated with no more than 100 retrieved images.
Since both these image datasets do not contain Japanese language, we download the Japanese-to-English dictionary online\footnote{https://github.com/facebookresearch/MUSE\#ground-truth-bilingual-dictionaries}.
We select words in each dataset that overlap with our caption vocabulary, which results in 230 German-English pairs in BERGSMA dataset, 1,311  German-English pairs and 1,217 French-English pairs in MMID dataset, and 2,408 Japanese-English pairs.

\subsection{Experimental Setup}
For the multi-lingual caption model, we set the word embedding size and the hidden size of LSTM as 512.
Adam algorithm is applied to optimize the model with learning rate of 0.0001 and batch size of 128.
The caption model is trained up to 100 epochs and the best model is selected according to caption performance on the validation set.

We compare our approach with two baseline vision-based methods proposed in \cite{kiela2015visual,hewitt2018learning}, which measure the similarity of two sets of global visual features for bilingual lexicon induction:
\begin{enumerate}
	\item CNN-mean: taking the similarity score of the averaged feature of the two image sets.
	\item CNN-avgmax: taking the average of the maximum similarity scores of two image sets.
\end{enumerate}

We evaluate the word translation performance using MRR (mean-reciprocal rank) as follows:
\begin{equation}
\mathrm{MRR} = \frac{1}{N} \sum_{n=1}^{N} \mathrm{max}_{y_m \in D(x_n)} \frac{1}{rank(x_n, y_m)}
\end{equation}
where $D(x_n)$ is the groundtruth translated words for source word $x_n$, and $rank(x_n, y_m)$ denotes the rank of groundtruth word $y_m$ in the rank list of translation candidates.
We also measure the precision at K (P@K)  score, which is the proportion of source words whose groundtruth translations rank in the top K words.
We set K as 1, 5, 10 and 20.

\begin{table} \centering
	\caption{Image captioning performance of different languages on the COCO and STAIR dataset.}
	\label{tab:coco_caption_results}
	\begin{tabular}{c|c|ccc} \toprule
		language & model & BLEU4 & METEOR & CIDEr \\ \midrule
		\multirow{2}{*}{English} & mono & \textbf{32.29} & 26.01 & 102.36 \\
		& multi & 31.59 & \textbf{26.07} & \textbf{102.41} \\ \midrule
		\multirow{2}{*}{Japanese} & mono & \textbf{39.85} & 31.63 & \textbf{98.12} \\
		& multi & 39.81 & \textbf{31.70} & 97.81 \\ \bottomrule
	\end{tabular}
\end{table}

\begin{table*}
	\centering
	\caption{Performance of German to English word translation.}
	\label{tab:de_en_translation}
	\begin{tabular}{c|c|ccccc|ccccc} \toprule
		& & \multicolumn{5}{c|}{BERGSMA} & \multicolumn{5}{c}{MMID} \\ 
		& & MRR & P@1 & P@5 & P@10 & P@20 & MRR & P@1 & P@5 & P@10 & P@20 \\ \midrule
		\multirow{2}{*}{baselines} & CNN-mean & 0.650 & 55.6 & 75.2 & 82.7 & 89.3 & 0.262 & 19.9 & 33.3 & 37.6 & 42.4 \\
		& CNN-avgmax & 0.723 & 65.0 & 79.9 & 84.1 & 87.9 & 0.430 & 38.5 & 47.2 & 49.8 & 52.8 \\ \midrule
		& linguistic & 0.755 & 67.6 & 86.2 & 88.0 & 91.6 & 0.467 & 38.8 & 55.5 & 61.6 & 67.4 \\
		proposed model & visual & 0.762 & 69.2 & 84.6 & 89.7 & 91.1 & 0.400 & 31.5 & 48.1 & 54.7 & 60.3 \\
		& linguistic + visual & \textbf{0.819} & \textbf{76.6} & \textbf{86.9} & \textbf{91.6} & \textbf{93.5} & \textbf{0.529} & \textbf{45.2} & \textbf{62.2} & \textbf{68.2} & \textbf{72.3}
		\\ \bottomrule
	\end{tabular}
\end{table*}

\subsection{Evaluation of Multi-lingual Image Caption}
We first evaluate the captioning performance of the proposed multi-lingual caption model, which serves as the foundation stone for our bilingual lexicon induction method.

We compare the proposed multi-lingual caption model with the mono-lingual model, which consists of the same model structure, but is trained separately for each language.
Table~\ref{tab:multi30k_caption_results} presents the captioning results on the multi30k dataset, where all the languages are from the Latin family.
The multi-lingual caption model achieves comparable performance with mono-lingual model for data sufficient languages such as English and German,
and significantly outperforms the mono-lingual model for the data-scarce language French with absolute 3.22 gains on the CIDEr metric.
For languages with distinctive grammar structures such as English and Japanese, the multi-lingual model is also on par with the mono-lingual model as shown in Table~\ref{tab:coco_caption_results}.
To be noted, the multi-lingual model contains about twice less of parameters than the independent mono-lingual models, which is more computation efficient.

\begin{figure}
	\begin{center}
		\includegraphics[width=1\linewidth]{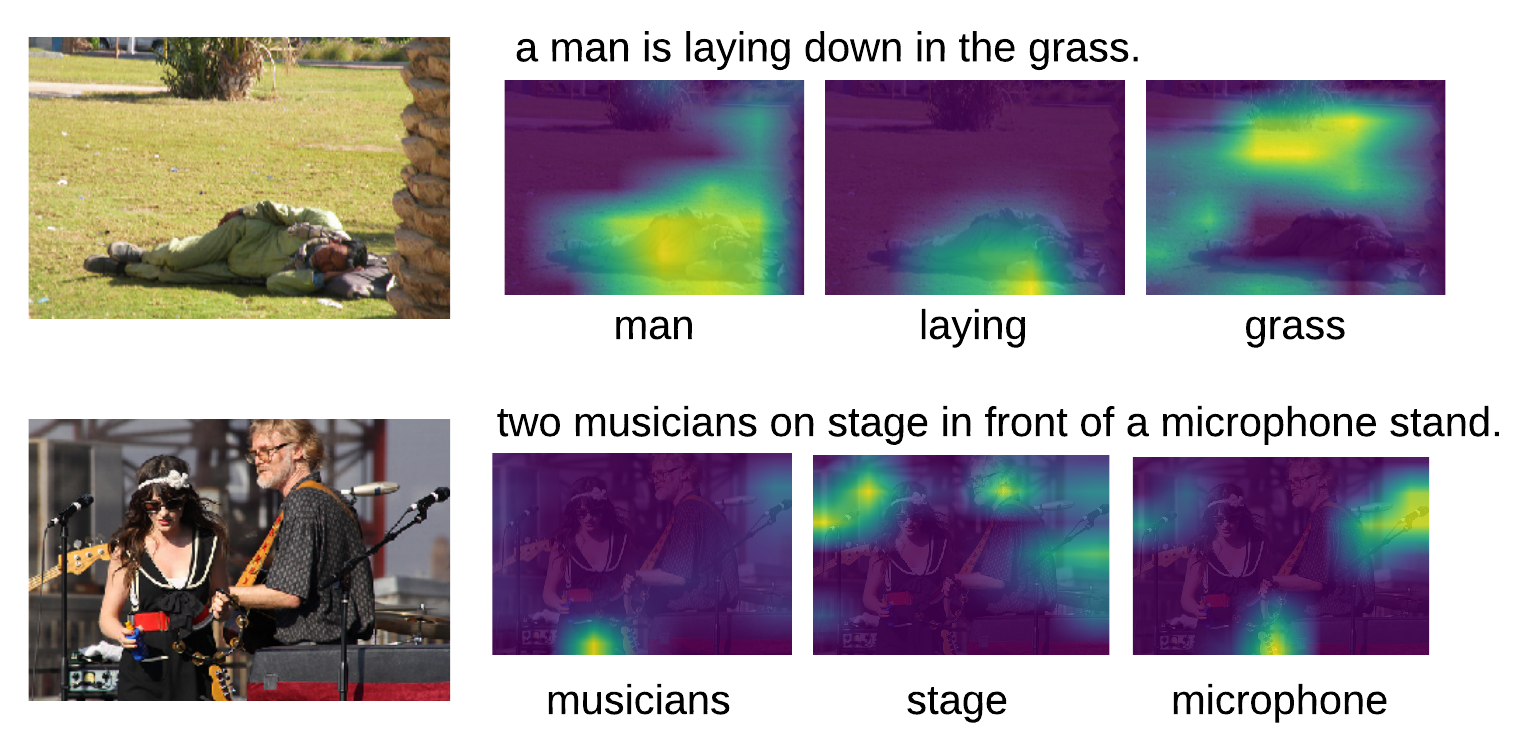}
	\end{center}
	\caption{Visual groundings learned from the caption model. } 
	\label{fig:attn_visualize}
\end{figure}

We visualize the learned visual groundings from the multi-lingual caption model in Figure~\ref{fig:attn_visualize}.
Though there is certain mistakes such as `musicians' in the bottom image, most of the words are grounded well with correct objects or scenes, and thus can obtain more salient visual features.

\subsection{Evaluation of Bilingual Lexicon Induction}

\begin{table}
	\caption{German-to-English word translation examples. `de' is the source German word and `en' is the groundtruth target English word. The `rank' denotes the position of the groundtruth target word in the candidate ranking list. The `top3 translation' presents the top 3 translated words of the source word by our system. }
	\label{tab:de_en_example}
	\begin{tabular}{c|ccc|l} \toprule
		pos & de & en & rank & top3 translation \\ \midrule
		\multirow{2}{*}{noun} & telefon & phone & 1 & phone, pay, telephone \\
		& gebiet & area & 10 & village, trees, desert \\ \midrule
		\multirow{2}{*}{verb} & sieht & looks & 1 & looks, stands, look \\
		& machen & make & 4 & do, take, performs \\ \midrule
		\multirow{2}{*}{adj} & roten & red & 1 & red, orange, pink \\
		& wenigen & few & 3 & many, several, few, \\ \midrule
		\multirow{2}{*}{adv} & sehr & very & 1 & very, large, huge \\
		& knapp & barely & 65 & pretty, short, ballet \\ \midrule
		\multirow{2}{*}{prep} & f$\ddot{u}$r & for & 1 & for, as, hold \\
		& oberhalb & above & 7 & below, near, within \\ \midrule
		\multirow{2}{*}{num} & einer & a, one & 1 & a, the, one \\
		& zehn & ten & 5 & seven, eight, six \\ \bottomrule
	\end{tabular}
\end{table}

We induce the linguistic features and localized visual features from the multi-lingual caption model for word translation from the source to target languages.
Table~\ref{tab:de_en_translation} presents the German-to-English word translation performance of the proposed features.
In the BERGSMA dataset, the visual features achieve better translation results than the linguistic features while they are inferior to the linguistic features in the MMID dataset.
This is because the vocabulary in BERGSMA dataset mainly consists of nouns, but the parts-of-speech is more diverse in the MMID dataset.
The visual features contribute most to translate concrete noun words, while the linguistic features are beneficial to other abstract words. 
The fusion of the two features performs best for word translation, which demonstrates that the two features are complementary with each other.

We also compare our approach with previous state-of-the-art vision-based methods in Table~\ref{tab:de_en_translation}.
Since our visual feature is the averaged representation, it is fair to compare with the CNN-mean baseline method where the only difference lies in the feature rather than similarity measurement.
The localized features perform substantially better than the global image features which demonstrate the effectiveness of the attention learned from the caption model.
The combination of visual and linguistic features also significantly improves the state-of-the-art visual-based CNN-avgmax method with 11.6\% and 6.7\% absolute gains on P@1 on the BERGSMA and MMID dataset respectively.

\begin{table}
	\centering
	\caption{Comparison of the image captioning models' impact on the bilingual lexicon induction. The acronym L is for linguistic and V is for the visual feature.}
	\label{tab:de_en_vanilla_caption}
	\begin{tabular}{c|c|ccccc} \toprule
		feat & \begin{tabular}[c]{@{}c@{}}caption\\ model\end{tabular} & MRR & P@1 & P@5 & P@10 & P@20 \\ \midrule
		\multirow{2}{*}{L} & mp & 0.437 & 35.5 & 52.9 & 58.8 & 66.4 \\
		& attn & \textbf{0.467} & \textbf{38.8} & \textbf{55.5} & \textbf{61.6} & \textbf{67.4}\\ \midrule
		\multirow{2}{*}{V} & mp & 0.367 & 28.8 & 45.0 & 50.2 & 57.1 \\
		& attn & \textbf{0.400} & \textbf{31.5} & \textbf{48.1} & \textbf{54.7} & \textbf{60.3} \\ \midrule
		\multirow{2}{*}{\begin{tabular}[c]{@{}c@{}}L+V\end{tabular}} & mp & 0.490 & 40.4 & 58.6 & 65.4 & 71.7 \\
		& attn & \textbf{0.529} & \textbf{45.2} & \textbf{62.2} & \textbf{68.2} & \textbf{72.3} \\ \bottomrule
	\end{tabular}
\end{table}

\begin{figure*} \centering
	\subfigure[MRR performance.] { \centering
		\label{fig:de_en_pos_proposed}
		\includegraphics[width=0.45\linewidth]{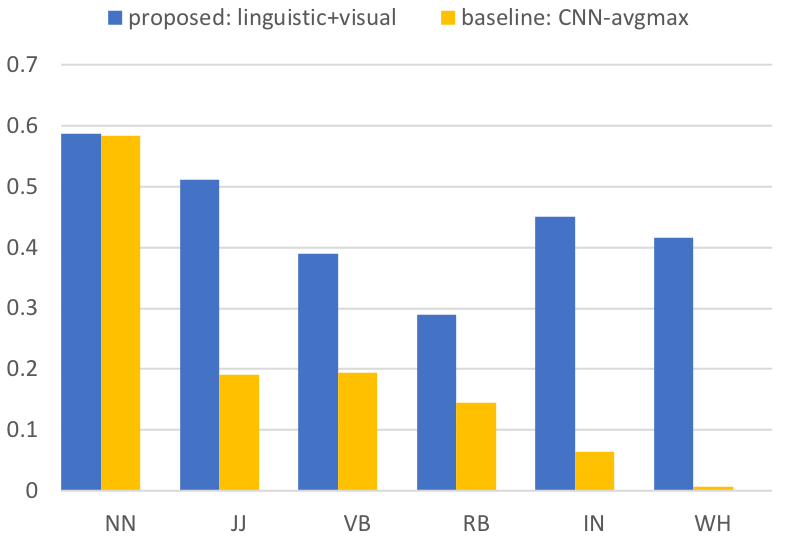}
	} 
	\subfigure[P@10 performance.] { \centering
		\label{fig:de_en_pos_baseline}
		\includegraphics[width=0.45\linewidth]{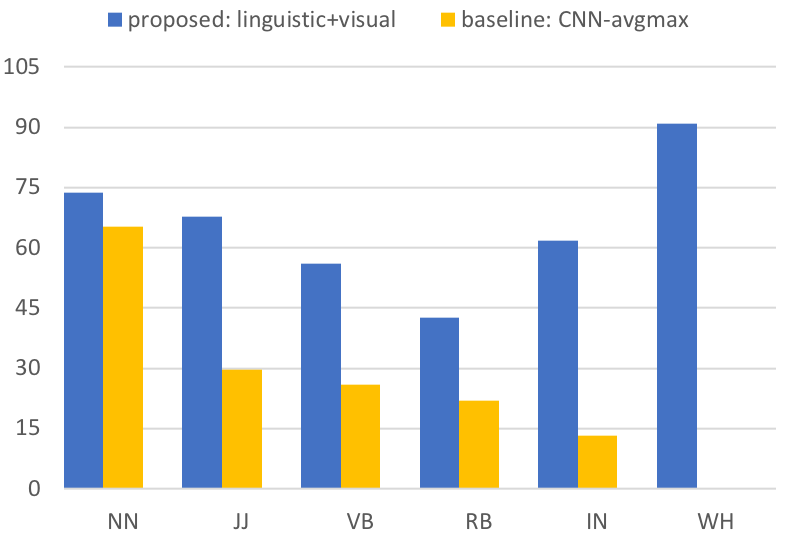}
	}
	\caption{Performance comparison of German-to-English word translation on the MMID dataset. The word translation performance is broken down by part-of-speech labels.}
	\label{fig:de_en_pos_results}
\end{figure*}

In Figure~\ref{fig:de_en_pos_results}, we present the word translation performance for different POS (part-of-speech) labels.
We assign the POS label for words in different languages according to their translations in English.
We can see that the previous state-of-the-art vision-based approach contributes mostly to noun words which are most visual-relevant, while generates poor translations for other part-of-speech words.
Our approach, however, substantially improves the translation performance for all part-of-speech classes.
For concrete words such as nouns and adjectives, the localized visual features produce better representation than previous global visual features;
and for other part-of-speech words, the linguistic features, which are learned with sentence context, are effective to complement the visual features.
The fusion of the linguistic and localized visual features in our approach leads to significant performance improvement over the state-of-the-art baseline method for all types of POS classes.

Some correct and incorrect translation examples for different POS classes are shown in Table~\ref{tab:de_en_example}.
The visual-relevant concrete words are easier to translate such as `phone' and `red'.
But our approach still generates reasonable results for abstract words such as `area'  and functional words such as `for' due to the fusion of visual and sentence contexts.

We also evaluate the influence of different image captioning structures on the bilingual lexicon induction. 
We compare our attention model (`attn') with the vanilla show-tell model (`mp') \cite{vinyals2017show}, which applies mean pooling over spatial image features to generate captions and achieves inferior caption performance to the attention model.
Table~\ref{tab:de_en_vanilla_caption} shows the word translation performance of the two caption models.
The attention model with better caption performance also induces better linguistic and localized visual features for bilingual lexicon induction.
Nevertheless, the show-tell model still outperforms the previous vision-based methods in Table~\ref{tab:de_en_translation}.

\begin{table}
	\caption{Performance of French to English word translation on the MMID dataset.}
	\label{tab:fr_en_results}
	\begin{tabular}{c|ccccc} \toprule
		& MRR & P@1 & P@5 & P@10 & P@20 \\ \midrule
		CNN-mean & 0.301 & 22.8 & 37.1 & 43.1 & 48.6 \\
		CNN-avgmax & 0.474 & 41.9 & 52.6 & 55.4 & 59.1
		\\ \midrule
		linguistic & 0.376 & 29.3 & 47.0 & 52.6 & 58.9 \\
		visual & 0.387 & 31.1 & 46.7 & 52.1 & 58.9 \\
		linguistic+visual & \textbf{0.494} & \textbf{42.0} & \textbf{57.1} & \textbf{62.9} & \textbf{69.4} \\ \bottomrule
	\end{tabular}
\end{table}

\begin{table}
	\caption{Performance of Japanese to English word translation.}
	\label{tab:jp_en_results}
	\begin{tabular}{c|ccccc} \toprule
		& MRR & P@1 & P@5 & P@10 & P@20 \\ \midrule
		linguistic & 0.290 & 22.1 & 35.8 & 42.0 & 49.5 \\
		visual & 0.419 & 34.2 & 50.4 & 56.3 & 61.1 \\
		linguistic+visual & \textbf{0.469} & \textbf{38.3} & \textbf{56.9} & \textbf{62.9} & \textbf{68.1} \\ \bottomrule
	\end{tabular}
\end{table}

\subsection{Generalization to Diverse Language Pairs}
Beside German-to-English word translation, we expand our approach to other languages including French and Japanese which is more distant from English.

The French-to-English word translation performance is presented in Table~\ref{tab:fr_en_results}.
To be noted, the training data of the French captions is five times less than German captions, which makes French-to-English word translation performance less competitive with German-to-English. 
But similarly, the fusion of linguistic and visual features achieves the best performance, which has boosted the baseline methods with 4.2\% relative gains on the MRR metric and 17.4\% relative improvements on the P@20 metric.

Table~\ref{tab:jp_en_results} shows the Japanese-to-English word translation performance.
Since the language structures of Japanese and English are quite different, the linguistic features learned from the multi-lingual caption model are less effective but still can benefit the visual features to improve the translation quality.
The results on multiple diverse language pairs further demonstrate the generalization of our approach for different languages.

\section{Conclusion}
In this paper,  we address the problem of bilingual lexicon induction without reliance on parallel corpora. 
Based on the experience that we humans can understand words better when they are within the context and can learn word translations with external world (e.g. images) as pivot, we propose a new vision-based approach to induce bilingual lexicon with images and their associated sentences.  
We build a multi-lingual caption model from multiple mono-lingual multimodal data to map words in different languages into joint spaces. 
Two types of word representation, linguistic features and localized visual features, are induced from the caption model. 
The two types of features are complementary for word translation.
Experimental results on multiple language pairs demonstrate the effectiveness of our proposed method, which leads to significant performance improvement over the state-of-the-art vision-based approaches for all types of part-of-speech.
In the future, we will further expand the vision-pivot approaches for zero-resource machine translation without parallel sentences.

\section{ Acknowledgments}
This work was supported by National Natural Science Foundation of China under Grant No. 61772535, National Key Research and Development Plan under Grant No. 2016YFB1001202 and Research Foundation of Beijing Municipal Science \& Technology Commission under Grant No. Z181100008918002.

\bibliography{reference}
\bibliographystyle{aaai}

\end{document}